# Advanced Efficient Strategy for Detection of Dark Objects Based on Spiking Network with Multi-Box Detection.

Munawar Ali*, Baoqun Yin, Hazrat Bilal, Aakash Kumar, Ali Muhammad, Avinash Rohra

*Department of Automation, University of Science and Technology of China, Hefei, Anhui 230026, People's Republic of China*

*Corresponding authors:

 E-mail address: alimunawar@mail.ustc.edu.cn (A. Munawar);

**Abstract**

Several deep learning algorithms have shown amazing performance for existing object detection tasks, but recognizing darker objects is the largest challenge. Moreover, those techniques struggled to detect or had a slow recognition rate, resulting in significant performance losses. As a result, an improved and accurate detection approach is required to address the above difficulty. The whole study proposes a combination of spiked and normal convolution layers as an energy-efficient and reliable object detector model. The proposed model is split into two sections. The first section is developed as a feature extractor, which utilizes pre-trained VGG16, and the second section of the proposal structure is the combination of spiked and normal Convolutional layers to detect the bounding boxes of images. We drew a pre-trained model for classifying detected objects. With state of the art Python libraries, spike layers can be trained efficiently. The proposed spike convolutional object detector (SCOD) has been evaluated on VOC and Ex-Dark datasets. SCOD reached 66.01% and 41.25% mAP for detecting 20 different objects in the VOC-12 and 12 objects in the Ex-Dark dataset. SCOD uses 14 Giga FLOPS for its forward path calculations. Experimental results indicated superior performance compared to Tiny YOLO, Spike YOLO, YOLO-LITE, Tinier YOLO and Center of loc+Xception based on mAP for the VOC dataset.

**Keywords:** Deep Learning, Object detection, Spiking Neural Network, Convolutional layer, single shot multi-box detector (SSD), VGG-16.

**Declaration of Competing Interest**

   On behalf of all authors, the corresponding author states that there is no conflict of interest.

# 1. Introduction

With the development of new technologies, gathering pictures and storing them on a local drive is getting easier. A recent review revealed that around 2016, smartphone and camera device consumers captured approximately 1.1 trillion photographs [1]. With the increasing availability and versatility of the pictures, the computer vision field has gained more attention. Object recognition is among the most complex issues in computer vision. Object detection involves two tasks in one framework: detecting the objects and classifying the detected objects [2]. A deep learning model is placed at the heart of each object detection framework. The utilization of deep learning has acquired more interest in object detection area due to the increase in the quality and accessibility of graphical computational power [3, 4] units. In recent years, two well-known categories for object detection with the names of "two-stage detection" and "one-stage detection" have emerged. Convolutional neural networks have been the most prominent deep learning techniques for the extraction of features in image detection techniques [5].

CNN can acquire a high-level image and robust visual features, which help to gather information from pictures for proposing candidate boxes for detecting objects. A few of the most affluent proposed networks for object detection are Regions with CNN features (RCNN)[5, 6]. The core concept of RCNN is the extraction of a series of object recommendations (object candidate box) through selective search [7, 8]. Each suggestion would subsequently be resampled towards a predetermined-sized picture as well as put through an ImageNet-trained CNN network. After that, linear Algorithms are utilized to forecast the existence of an image inside every section as well as to distinguish object classifications. You Only Look Once or YOLO in short is another network for object detection tasks [9, 10].The mixed-precision neural networks, which are energy-efficient, are utilized in addressing the issue which transmits the information towards the cars. The automotive infrastructure has been optimized for data storage via unmanned cars' mixed precision values with neural networks [11]. YOLO uses one neural network for the detection of objects in the full image. Instead of surveying pictures as one big structure, YOLO splits images into regions with expected bounding boxes as well as probability for every section at the same time. [12–14]. a single shot multibox detector (SSD) was another object detection framework that was subcategorized as a one-stage detection branch [15]. SSD's significant element is the development of multi-resolution and multi-reference recognition algorithms. All of the mentioned methods have focused on using different versions of the convolutional layer for object detection.

In this research we use spiking convolutional networks instead of conventional convolutional neural networks. Spiking neural networks were initially investigated as simulations of biological pattern recognition [16], wherein neurons communicate through spikes. Spiking neurons wait for information from the receptive field to get higher than specific thresholds and then fire the communication signal to other neurons. The use of spiking neural networks (SNN) typically helps to decrease the total neuron's existence in the network and make the network efficient based on how much space is needed for computation [17]. Despite the effective procedure of the SNN for reducing the computational power of deep learning models, there are also some drawbacks in SNN. One of the drawbacks is the sophisticated procedure for training the SNN. The use of SSNs for object detection has recently gained attention. One study in early 2020 combined the structure of YOLO with SNN for object detection tasks [18]. They concluded that Spiking YOLO uses around 280 times less power energy compared to the tiny YOLO. However, the results of their work based on the mean absolute precision (mAP) were not very good. They reached 51.83 mAP for object detection on the PASCAL VOC dataset.

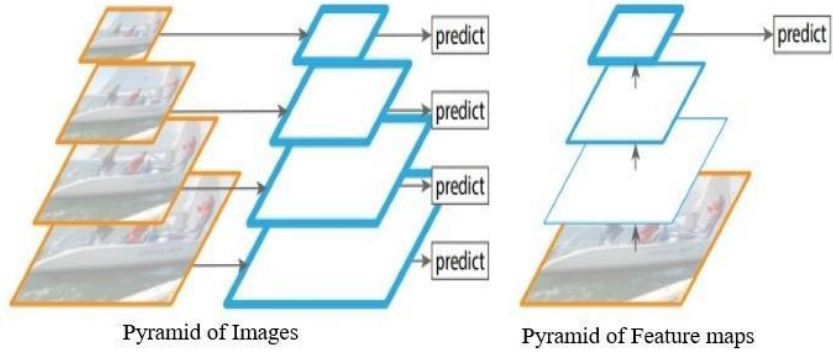

**Fig. 1** Structure of different layers of the pyramid CNN.

For the construction of high-level semantic feature patterns throughout all sizes, the best framework featuring lateral connections is constructed. The structures of such a network are shown in Fig.1. The layout has transversal connections which are used in feature pyramid networks (FPN)[19, 20] for the development of slightly elevated semantics throughout all stages. Despite the drawbacks of the YOLO model, it is considered as the quickest object identification algorithm with real-time capabilities. YOLO is constantly developing, and its performance is also improving [21]. Tiny-YOLO-V3 [21] and Tinier-YOLO [22]. They represent the most recent enhancements of YOLO, which is a relatively small model with relatively good accuracy. In summary, real-time object detection frameworks suffer from the trade-off between accuracy and the huge memory consumption process of training and testing. In this research we combined SSD with a spiking layer in order to not only increase the efficiency of energy consumption for the object detection framework but also to increase accuracy based on mAP. We used different versions of the Pascal [23] and Dark-Net [24] datasets for the evaluation of the proposed model to increase the accuracy, and we utilized the auxiliary path on the convolutional layer to see if the SNN layer works better than the ordinal convolutional layer. Finally, for more accurate analyses, we evaluate the overall results of the algorithm with different variants of YOLO and SSD. The experimental results indicate 66% mAP for detecting objects on VOC12 dataset and 40% mAP on the Dark-Net dataset.

## 2. Literature Survey

As discussed before, the use of convolutional layers for object detection is common. In recent years the use of deeper networks or the use of new loss functions has gained attention. In work by T-Y. Lin et al [25]. They proposed a pyramid network for recognizing entities at various sizes throughout images. Pyramid networks are individual layers that help to detect objects at different dimensions. Because a CNN automatically develops a feature pyramid using onward propagation, the FPN makes significant progress in recognizing features of various sizes. Different versions of RCNN have already been developed. Faster RCNN is the first end-to-end, and the first near-real time deep learning detector [26]. From R-CNN to Faster RCNN [27], the majority of an object identification system's separate blocks, such as object detection, extraction of features, boundary boxes prediction, and many more, have been progressively merged into a coherent, end-to-end development model. In recent works by T.Y. Lin et al [25]. They proposed a new loss function named "focal loss". Researchers stated that the high ground truth imbalanced data experienced throughout dense detection activation is the root problem. Throughout development, the focal calculation altered the conventional cross-entropy loss lost to ensure that the detection focused further on problematic, misclassified samples. Due to focal loss, single-stage detectors can attain the same precision as numerous detection techniques while maintaining extraordinarily rapid detection performance [25]. Faster RCNN is another version of RCNN that uses the original structure of the pyramid network for real-time object detection [27]. Faster RCNN is evaluated with the use of GPU and reported 33% detection at 15 frames per second (FPR). Other than using a proper model for object detection, the use of a proper dataset is also important. Object detection tasks require high-volume data with more class and diversity in the dataset. With the growth in the usage of more computational power, new datasets for object detection have been proposed. Some samples of these datasets are: PASCAL VOC Challenges [28] (e.g., VOC2012, VOC-07), ImageNet Large Scale Visual Recognition Challenge (e.g., ILSVRC2014) [29], MS-COCO Detection Challenge [30], etc. Another famous model for object detection is YOLO. The earliest YOLO model includes 1 GB of memory, which is relatively huge [21][15]. Since the original combination took up a lot of space, the recently developed version of YOLO has decreased this large volume in the work by FANG et al. [22]. They revised the YOLO-v3 model to decrease the original size while maintaining acceptable accuracy. This model includes six max pooling layers and seven convolutional layers, also including two detecting modes for identifying picture characteristics. They reported 65.7% mAP on the VOC007 dataset with a model size equal to 8.7 MB in another work by Huang et al [4][4]. They proposed the YOLO-LITE model for object detection. Their proposed structure is composed of 7 convolutional and 5 max pooling layers. They reported a mAP of 33.7% on the VOC07 dataset. Another attempt to minimize the original volume of the object detection model can be seen in fast YOLO [31]. This model is a simple version of the original YOLO with a 1*1 convolutional layer. Fast YOLO has 17.1 million parameters for training, and it is 3.3 times faster compared with the original YOLOv2. The spiking layer is the original structure of nature for firing neurons in a biological neural network. Kim et al. [18]. They combined the spiking strategy with the YOLO framework to decrease the size of the model while preserving the accuracy of detected objects. They reported a mAP of 51.8% on the VOC12 dataset. As seen from these previous works, the aim of these models is to reduce the testing run-time without increasing the mAP. The performance of different models is listed in Table 1.

**Table 1** Surveying different models for object detection with their mean average precision.

| Authors | Dataset | Model | mAP | Type |
|---|---|---|---|---|
| Jasper et al. [8] | VOC07 | RCNN | 58.5% | Not real time |
| he et al. [32] | VOC07 | Spatial pyramid pooling network | 59.2% | Not real time |
| girshick [33] | VOC07 | Fast RCNN | 70% | Not real time |
| Lin et al.[30] | COCO | Feature pyramid networks | 51.9% | Not real time |
| Joseph et al.[9] | VOC07 | YOLO | 63.4% | Not real time |
| Zhao et al. [34] | VOC07 | YOLO3 | 72.0% | Not real time |
| Bochkovskiy et al. [35] | MSCOCO | YOLO4 | 43.5% | Not real time |
| Liu et al. [15] | VOC07 | SSD | 76.8% | Not real time |
| Lin et al.[25] | COCO | Retina Net | 59.1% | Not real time |
| Li et al. [36] | COCO | Trident Net | 69.7% | Not real time |
| Kim et al. [18] | MSCOCO | Spiking YOLO | 26.2% | Real time |
| Haung et al. [4] | VOC12 | YOLO-LITE | 33.7% | Real time |
| FANG et al [22] | VOC07 | Tinier-YOLO | 65.7% | Real time |
| Zhang et al [37] | VOC07 | Tiny-YOLO | 53.0% | Real time |
| Ibrahem et al [38] | VOC07 | Faster RCNN+VGG16 | 50.3% | Real time |

Time and resources are other factors that make a presented object detection model valuable. Based on reviewed research for, mean average precision is the main aim of real-time frameworks for object detection. However, lowering the computational time also causes a decrease in the accuracy of the object detection framework, so we focused on this challenging task. In this article, we proposed a framework to minimize computational running time while preserving the accuracy of the predictions. We used a combination of the SSD and SNN models to overcome this challenge. The experimental results indicated a 66% mAP on the VOC-12 dataset which is higher than all reported real-time predictors thus far.

## 3. Material and Methods

This section explains a summary of the datasets and methodology of the work. First, we explain the VOC and Ex-dark datasets in detail. Then we specify the spiking layer and how it can be used to minimize the computational cost of the object detection model.

### 3.1 Pascal Visual Object Classes (VOC) Dataset

The PASCAL VOC dataset includes 20 parameter classes, such as vehicles, house-holds, animals, airplanes, bicycles, boats, buses, cars, motorbikes, trains, bottles, chairs, dining tables, potted plants, sofas, TV/monitors, birds, cats, cows, dogs, horses, sheep, and person. These categories can be summarized into animal categories, furniture categories, vehicles, and persons [39]. A sample of these pictures is shown in Fig. 2.

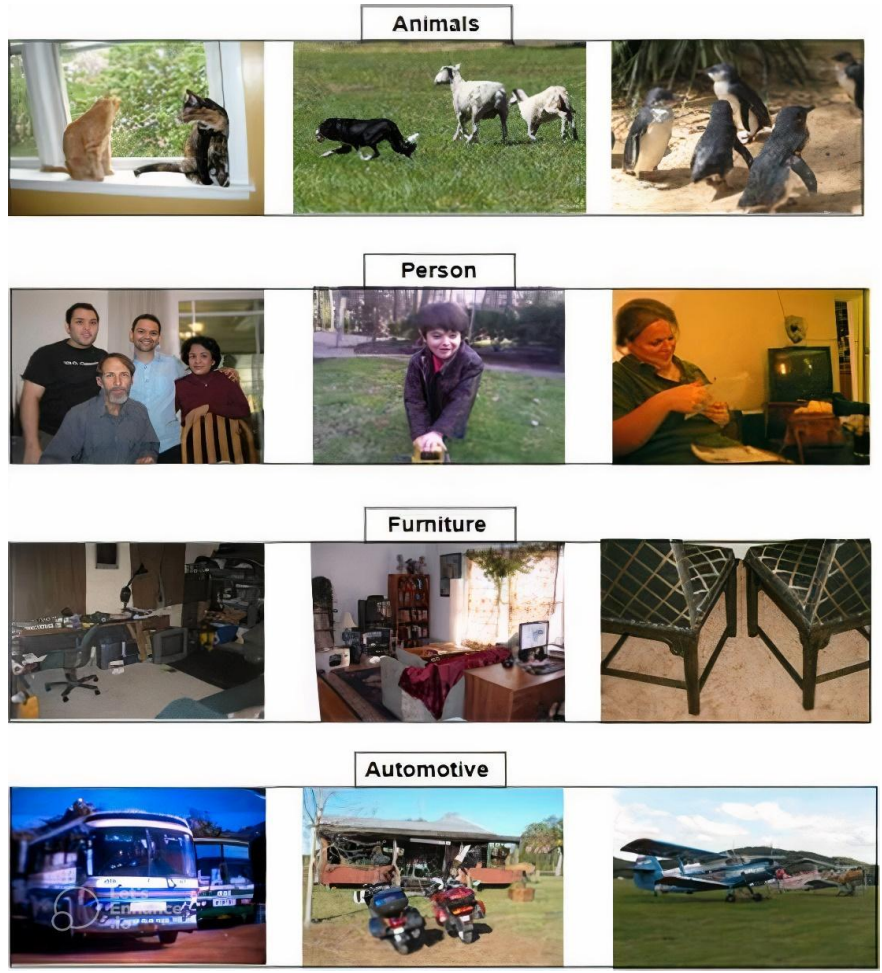

**Fig. 2** A demonstration of various categories of the VOC dataset.

VOC has different versions that were published in 2007 and 2012[40] [34]. VOC is a standard dataset for object detection that contains picture information with corresponding labels for bounding boxes and their labels. In this research we used the 2007 published VOC training set (VOC-07) and 2012 published (VOC-12) training and test sets. We developed our models on the VOC 07 and VOC 12 training sets and evaluated the proposed models on the VOC 12 test set. VOC 07 contains 9,963 images as training/validation images. VOC 12 contains 11,530 training/validation images.

### 3.2 Exclusively-Dark Dataset

Another dataset that is used in this research is the Ex-Dark dataset. This dataset contains information about 12 classes of objects such as boats, buses, cars, bicycles, bottles, cats, chairs, cups, sofas, dogs, tables, motorbikes, and persons [41]. The Ex-Dark dataset is collected in twilight situations, and most of the dataset's objects are unclear. A sample of these pictures is shown in Fig.3. Ex-Dark is indeed a set of 7,363 low-light photos ranging from extremely poor light to twilight. This dataset is similar to VOC 07, and it contains bounding boxes of the Ex-Dark dataset. The distribution of classes in the dataset is balanced, and it is between 8 to 10% for each class. The Ex-Dark dataset is utilized to evaluate the proposed model in improper situations. For the evaluation of the model, we split the entire dataset into 30% validation and 70% training for better results. Below down, Fig. 3 is representing the Ex-Dark dataset with correlated annotations.

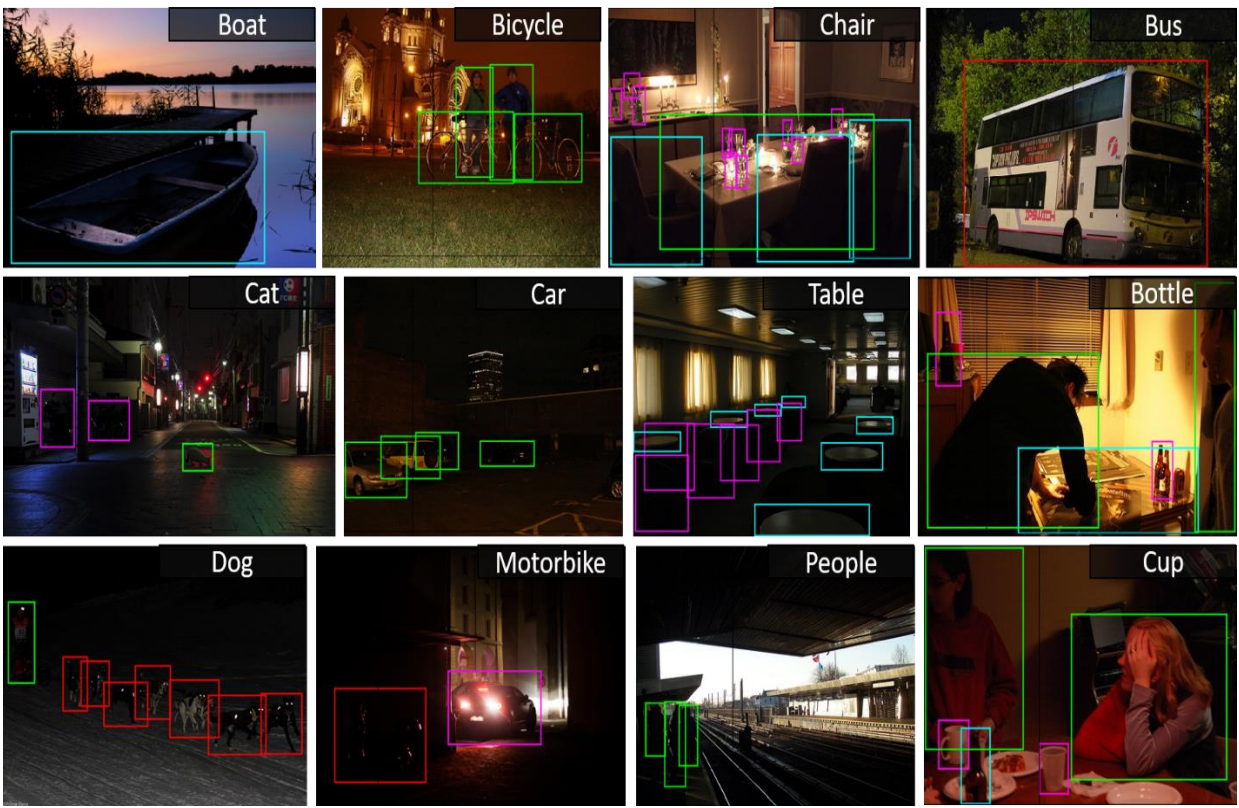

**Fig. 3** A demonstration of various categories of the Ex-Dark dataset with correlated annotations [41].

### 4 Methodology

As mentioned before, CNNs are the heart of many object detection models. The main problem with the CNN model is the large computational power and complexity of the network to be compiled in real-time applications [37]. CNN utilizes each connected layer's spatial information and shared weight to extract useful features. By increasing the depth of the model, more complex structures from input pictures, such as circles and curves are extracted [42]. Optimizing the CNN model is typically achieved by the backpropagation technique. The backpropagation equation updates the weight of the DNN model with the use of the chain rule for calculating the derivatives. Back-propagation formula is shown in Eq. (1):

$$W_{i,j}^{t+1} = W_{i,j}^{t} - lr * \left(\frac{\sigma y}{\sigma x}\left(\nabla\left(W_{i,j}^{t}\right)\right)\right) \qquad (1)$$

Where $W_{i,j}^{t}$ refers to the i and j locations of the trained weight at time t, and $lr$ refers to the learning rate of the optimizer. Each iteration depends on the CNN model's depth; gradients of the initial weights are calculated. As the depth of layers in the model increases for better feature extraction of the input images, the related computations for optimization of the model will also escalate. To specify the computational cost of the CNN model, we compute the overall floating-point operations (FLOPS). FLOPS would be the quantity of multiplying operations and accumulating dot product operations executed for each inference and per feedback [43]. The formula for the FLOPS calculation of a simple CNN model is shown in Eq. (2):

$$FLOPS_{CNN} = O2 * N * k2 * M \qquad (2)$$

Where M is the output channel, N represent input channels, the input map size is $1 \times 1$, kernel weight size represents by k, and O is the output size. Based on Eq. (2), each layer has its own computational complexity, and with increasing input size, the calculation complexity of the CNN model will increase. On the other hand, with decreasing input picture size of the model, the complexity of computations of our model is decreased, but that also affects the quality of the detected objects. In this research, to resolve this problem, we proposed an object detection model based on a combination of the spiking layer with a conventional convolutional layer to obtain better mean average precision.

### 4.1 Spiking Neural Network

A spiking neural network (SNN) employs signal trains, which are a sequence of events, to deliver info messages among neurons. In spiking network input and output are usually decoded. The leaky integrate and fire (LIF) model base on SNNs were utilized for energy-efficient space-time and spatiotemporal data processing. The LIF-SNN advantages via event-driven computation are owing to realistic neural dynamics as well as compactness; nonetheless, it typically suffers the disgrace of poor low efficiency [44]. The strategy behind the spiking layers is explained as follows. The integrate-and-fire neurons accumulate input Xi into a membrane potential Pm. The equation of this structure is shown in Eq. (3):

$$P_{mem,j}^{l}(t+1) = P_{mem,j}^{l}(t) - X_{j}^{l}(t) - V_{th} * \theta_{j}^{l}(t) \qquad (3)$$

Where $\theta_{j}^{l}$ refers to spike signals and $V_{th}$ refers to the threshold voltage. $X_{j}^{l}(t)$ feedback from the $j_{th}$ neuron inside the corresponding layer. A spike (θ) is formed once the integral value $P_{mem}$ crosses the threshold voltage $V_{th}$. The formula for such calculations is shown in Eq. (4):

$$\theta_{j}^{l-1}(t) = U * (P_{mem,j}^{l}(t) - V_{th}) \qquad (4)$$

Here U represents a step input unit. Eq. (3) and (4) signify the idea behind the spiking strategy. In spiking convolutional structure instead of using the spiking strategy on an individual neuron as input, spiking convolutions try to focus on using this structure on a shared weight window as input signals. For better performance by Utilizing spiking layer inside of the object detection model, a combination between the spiking layer and convolutional layer is implemented [45]. In our research we used a combination of convolutional layers and spiking layers with the name of spiking convolutions. The output of such a layer is shown in Eq. (5):

$$P_{mem,j}^{l}(t+1) = P_{mem,j}^{l}(t) + \sum W_{i,j}(t) * \delta\left(t - t_{f}^{l}(j)\right) \qquad (5)$$

Where $P_{mem,j}^l$ refers to the membrane potential, $t_f^l(j)$ corresponds towards the j$^{th}$ neuron activation duration within layer l, $W_{i,j}(t)$ represents the strength of the jth synapse in the spiking convolution $W_{i,j}$ represents the weighted connection to the input windows of shared weights structure, and δ refers to the Kronecker delta function. A Kronecker delta function is generally defined as a relationship between two parameters, which are typically nonzero values. Whereas if parameters are the same, the method returns 1; elsewhere, it returns 0. In a spiking layer, a neuron's current state is defined as its membrane potential. The value is changing through the process of learning. Unlike CNNs, SNNs convey data using the precision (spatiotemporal) of spike trains made up of a succession of spikes (spatially separated), instead of an actual amount (continuous). SNNs, like neural networks in nature, use feature representations in transmitting data. Furthermore, once spikes get generated, the spiking neurons incorporate information together into membrane potential and create (fire) events whenever the membrane potential surpasses a particular limit, enabling event-driven computing [46]. SNNs provide hegemonic consumption and have become the preferable neural network in artificial neural structures due to the dense quality of spike occurrences and event-driven computing. The strategy for updating the spiking layers is mentioned in Eq. (6):

$$W_{i,j}^{t+1} = W_{i,j}^t - lr * \left(S_d(t) - S_j(t) * S_i(t)\right) \qquad (6)$$

Where $W_{i,j}^t$ is the efficacy of the synaptic coupling from receptive input windows, $S_d(t)$ is the target (reference) spike train, $S_i(t)$ is the low-pass filtered spike train input, $S_j(t)$ is the spike train output, $lr$ represents the learning rate. By changing the formula for calculating the optimizer weight and how the model works, the computational complexity also diminishes. The formula for computational complexity is shown in Eq. (7):

$$FLOPS_{SNN} = O2 * N * k2 * M * S_a \qquad (7)$$

As illustrated in Eq. (7), spiking activity, $S_a$ is a very important factor that minimizes the FLOPS value by 0.0000001 to 0.00000001 of the original layers based on the activity of each spiking layer [43]. For the calculation of the energy of each spiking layer the following formula has been suggested by Panda et al as shown in Eq. (8):

$$E_{SNN} = \sum_{i=1}^{n} FLOPS_{SNN} * 0.1 * T \qquad (8)$$

Wherein N represents the amount of computing layers. The performance estimation of SNN involves the latencies imposed because the rate-coded input spikes trains must be conveyed across a number of times to provide the ultimate prediction accuracy. So, in summary estimated power used by the spikes layers in the structure of memorizing input tensor is between 0.00000001 and 0.00000001 less than $N$ original convolutional layer.

## 4.2 Proposed Model

By surveying CNN and SNN structures in this research, we investigate the problem and restriction of each of these structures. In CNNs, the main problem is the computational complexity and number of necessary FLOPS for optimizing the CNNs. On the other hand, the SNN problem is low performance and a hard optimizing process due to the harmonic times for updating each of the spiking signals [47]. In this research, by acknowledging the above mentioned problems, we propose a combination of a convolutional layer and spiking layers to solve the problem of poor performance by spiking layers and high computational structure by CNN.

We introduce a deep neural network technique for recognizing objects in pictures. This method decomposes an image of the output matrix of bounding boxes together into a set of default boxes with proportionate variation and weights per position of the feature map. Throughout classification, a model computes values representing its presence of all categories inside this default box and also modifies those boxes that best reflect the shapes of the object. Additionally, the algorithm manages images of varying scales intuitively via integrating predictions from several extracted features using different resolutions. This model is straightforward in comparison to many other approaches that need classifiers thus it prevents formulating creation and associated feature resampling or pixel phases and integrates the whole process in a particular network. We proposed a new object detection model which split into two sections. The first section is composed of a backbone network that uses the pretrained strategy to execute features from the input pictures and various famous pretraining models, such as the VGG16 network for available candidates for feature extraction of the input pictures [48]. Among these feature extractors due to the cause of the simple and shallow model, we chose to work with VGG16 and Inception as the candidates for feature extraction. Our proposed model for object detection is shown in Fig. 4.

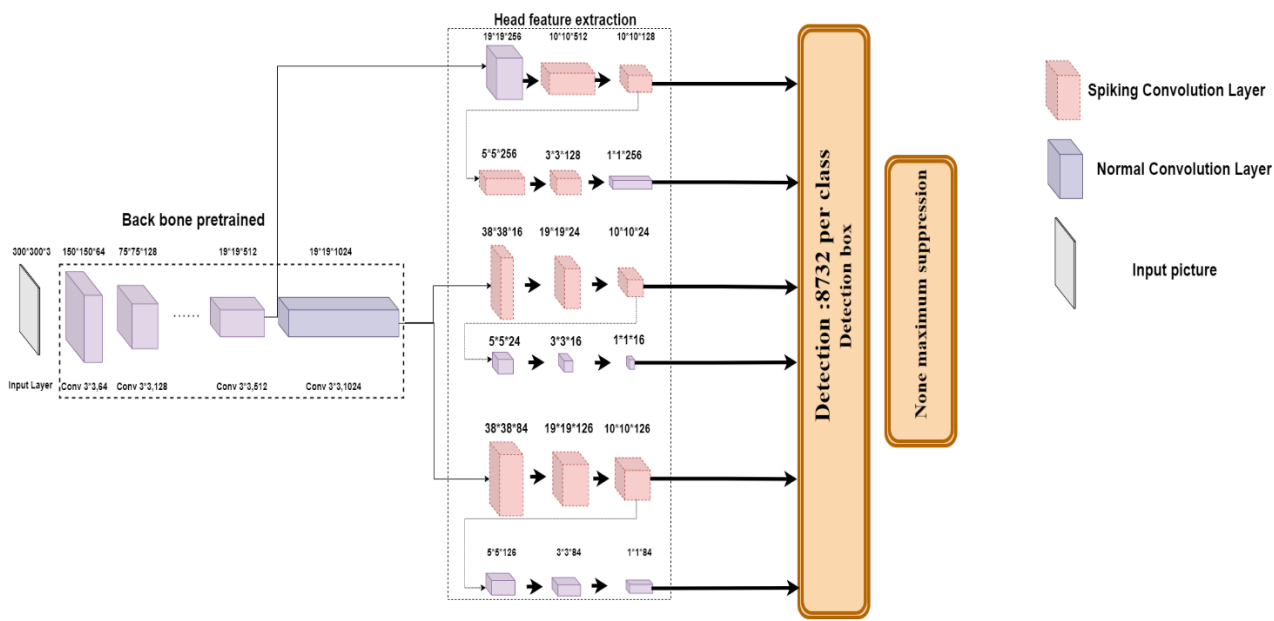

**Fig. 4** Proposed model for object detection

The second section or head is utilized for classification and bounding box prediction. In this section, we used spiking convolutional layers with a combination of spiking layers to solve the problem of high computational power. The head of the proposed model consisted of three main paths that use the spiking convolutional layers as the main path to detect the bounding 7 boxes and a convolutional to categorize the identified objects and decrease the effect of the spiking layer for the downside effect of the spiking layers. Our major goal is to eliminate the high computational needs for the convolutional layers in the model; therefore, based on Eq. (2) the size of the kernel, size of the input channels, and output size of the convolutional layer decrease through this proposed model, to minimize the necessary computing FLOPS. Furthermore, to better conclude the auxiliary routes in the head, we utilized a convolutional layer with low input, output, and kernel size.

The proposed structure is shown in Fig.4. The backbone of the model is part of the VGG16 network as the network escalates then the depth of the model increases. In the head part each of the auxiliary paths contains another three spiking layers with conventional convolutional layers. The output of these convolutional layers is fed into the detection bounding boxes. Detected bounding boxes usually have a size of 10*10 or 1*1 as the kernel size and number of filters vary between various sizes, such as 256, 128, 16, and 84. Finally, 8732 bounding boxes are predicted for each of the classes. In order to find the best bounding boxes for each object. The proposed structure is surveyed by the cross-categorical loss function for classification and location loss for specifying the best candidate as windows for specifying the best size for bounding boxes [49]. To ensure that classification and predicted bounding boxes were trained perfectly on the process, we calculated mean average precision (mAP) for better accuracy of the bounding boxes.

## 5 Experimental Results

In this research a new model with the combination of spiking layer and a conventional convolutional neural network is presented for object detection. The proposed model is evaluated on the VOC-12 and Ex-Dark datasets. In addition to assessing the suggested architecture, we used PyTorch libraries to create and use spiking convolutional layers [50]. To train the model, we used the T4 GPU with multi-thread computing to train the proposed model. The result of the training and evaluation of the model on VOC 12 is shown in Fig. 5.

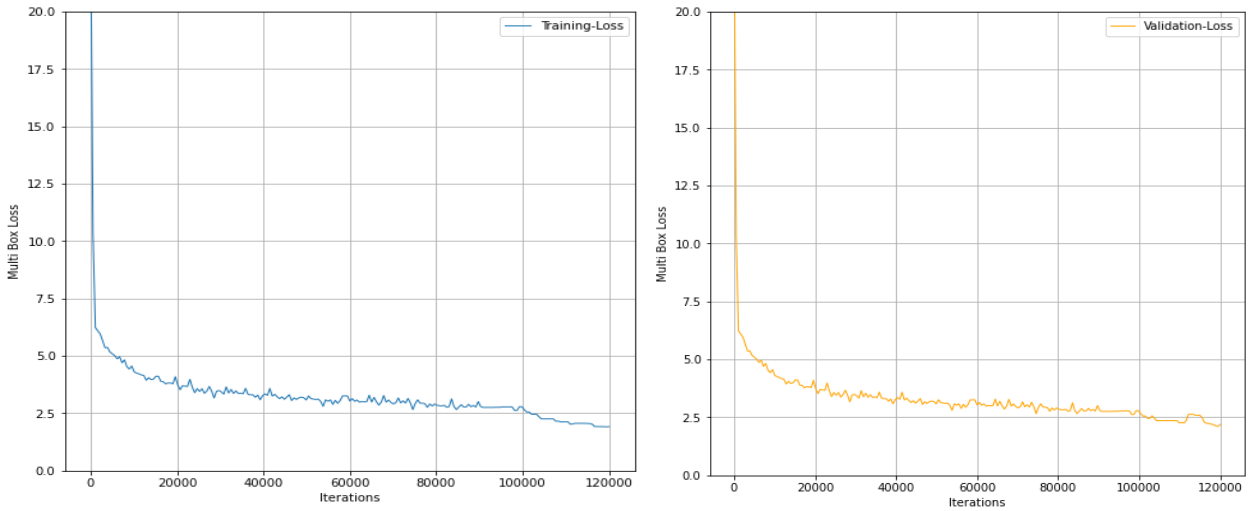

**Fig. 5** Training loss and validation loss of the Multi-Box detection on the VOC-12 dataset.

These results properly evaluated on VOC-12 training set and validation set are also compatible with each other and show no overfitting issue in the training process. The result of evaluating the model based on mean average precision (mAP) is shown in Table 2. Our proposed model detected objects in various classes with different mAP. The best criteria are achieved on the cat class and the worst performance is achieved in the bottle class. Although important innovations in object detection have been mostly developed once after the next, they clearly cope with bright pictures while severely deficient in low light. We conclude that it must be attributable to a shortage of available datasets to enable and evaluate development throughout this subject. These are the mean average precision results on the VOC-12 dataset. We calculated the mean average precision (mAP) to analyze the object detection algorithms for the best performance. The mAP computes a score by comparing the ground-truth bounding box to the identified box.

**Table 2** Results of evaluating the proposed model on the VOC-12 dataset

| VOC Classes | Average Precision. AP (%) |
|---|---|
| Airplane | 70.26 |
| Bicycle | 75.35 |
| Bird | 65.36 |
| Boat | 54.58 |
| Bottle | 28.20 |
| Bus | 73.15 |
| Car | 79.68 |
| Cat | 83.68 |
| Chair | 39.76 |
| Cow | 68.54 |
| Dining table | 64.15 |
| Dog | 81.47 |
| House | 80.58 |
| Motor bicycle | 74.48 |
| Person | 70.07 |
| Potted planet | 31.74 |
| Sheep | 66.60 |
| Sofa | 68.02 |
| Train | 77.46 |
| TV | 66.12 |
| **Overall** | **66.01** |

In this paper, the more accurate the result, the further exact the model's detection procedure, we discuss the predictions of the model and model accuracy. We also deployed the Python libraries module to compute those measures. We now show how accuracy and memory are utilized to determine the mean average precision (mAP) in Fig. 6.

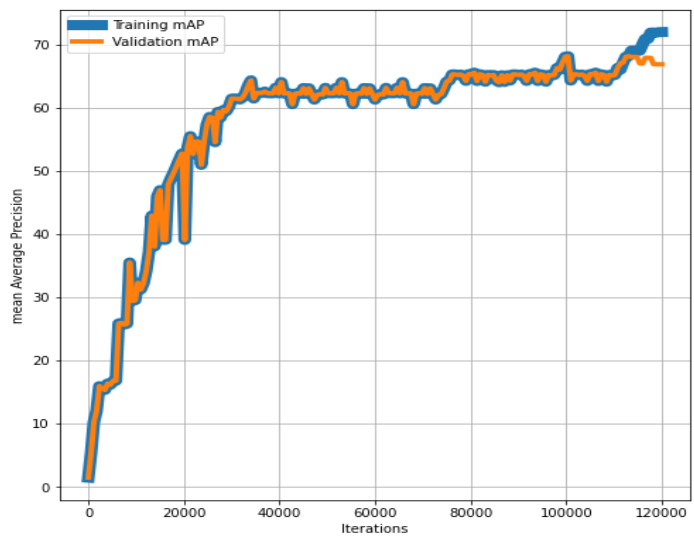

**Fig. 6** Our Model Mean Average Precision mAP (Training and Validation) on the VOC dataset results.

In the case of the VOC dataset, the proposed model is trained for the same number of iterations, but the loss converges to a lesser value than the Ex-Dark dataset. The result of mAP reached 66.01% for validation, which is higher than the training structure. Below are the results of testing the proposed model on different classes on the VOC-12 dataset shown in Fig. 7.

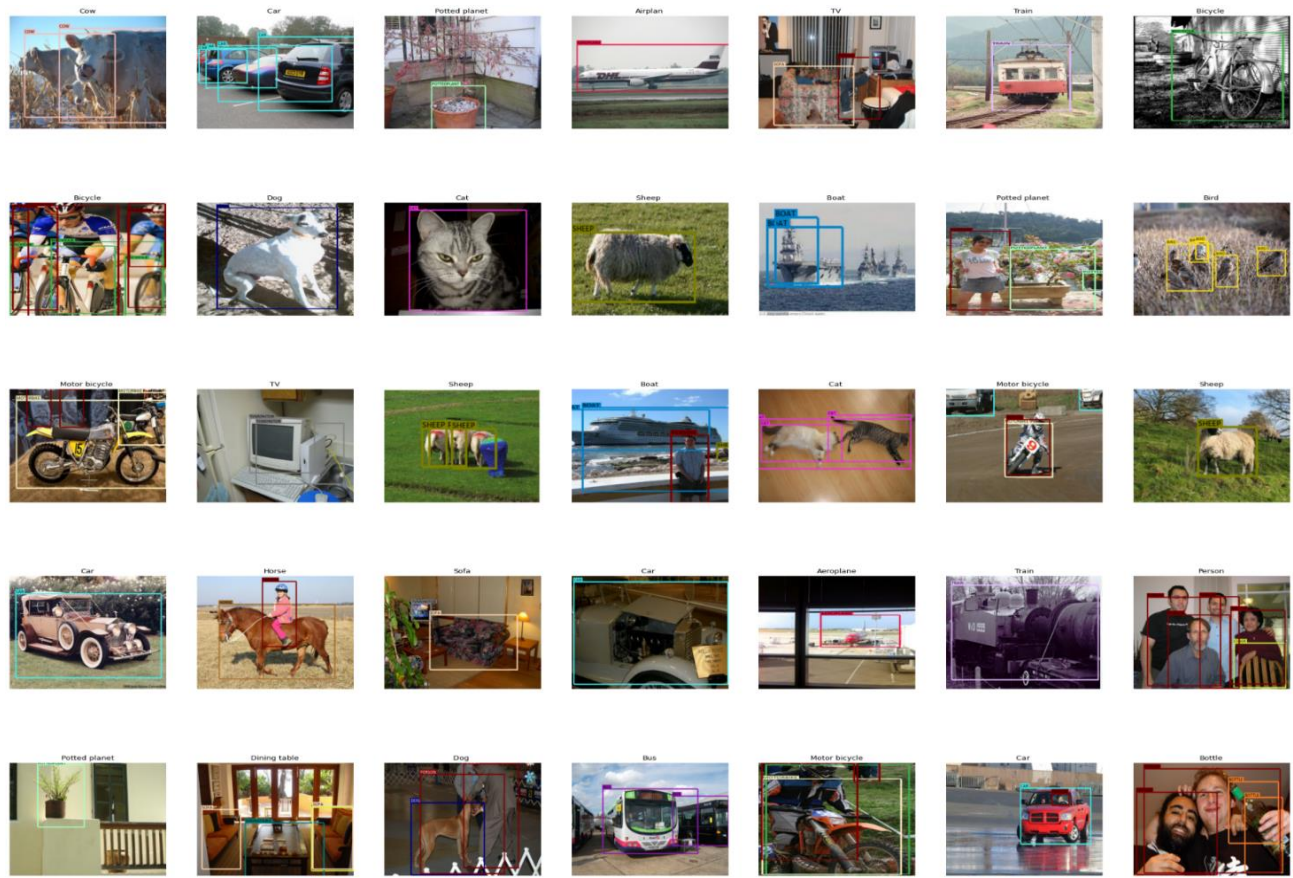

**Fig. 7** Result of testing the proposed model on different classes on the VOC-12 dataset

On the other hand, due to the twilight condition for collecting pictures, the proposed model has reported lower performance for detecting objects in the dark. The results of detecting objects in the Ex-Dark dataset have been shown in Table 3. As it has shown the performance for detecting the objects in the VOC dataset is higher than detected objects in the Ex-Dark dataset. The reason behind a certain disparity in efficiency can be seen in the process of evaluation and how the model is going to detect the edges of the objects. In dark and twilight situations, the boundary between objects in the pictures is not clear, and this fact will affect the process of detecting images in the pictures. Thus, it is clear that the process of detecting objects in the Ex-Dark dataset is harder than that in the VOC dataset. To check the results of the testing models on various pictures in different positions, we used a test set of VOC, Ex-Dark, and wild pictures. The result of detecting objects in these pictures is shown in Fig. 10. One of the main reasons to use the proposed model is to decrease the computational cost for the forward path in order to solve the problem of dark objects detection.

Table 3 Result of evaluating proposed model on Ex-Dark dataset

| Ex-Dark Classes | Mean Average Precision mAP (%) |
|---|---|
| Bus | 60 |
| Car | 56 |
| Cat | 54 |
| Dog | 20 |
| Table | 11 |
| People | 56 |
| Bottle | 13 |
| Chair | 39 |
| Motor bicycle | 53 |
| Cup | 40 |
| Boat | 45 |
| Bicycle | 48 |
| **Overall** | **41.25** |

Based on Eq. (7) for calculations of the spiking layer and conventional convolutional layer with a kernel size of 3×3, an input channel size of 128, an out channel size of 256, and an output channel size of 512×10×10 and $S_A$ equal to 0.00000001, the number of FLOPS is equal to 0.069 Gigaflops. With the same formulation, the proposed structure has almost 15 Giga FLOPS, most of them belong to the backbone of the proposed structure. Below down, our model training loss and validation loss of multi-box on exclusively-dark dataset results are shown in Fig.8.

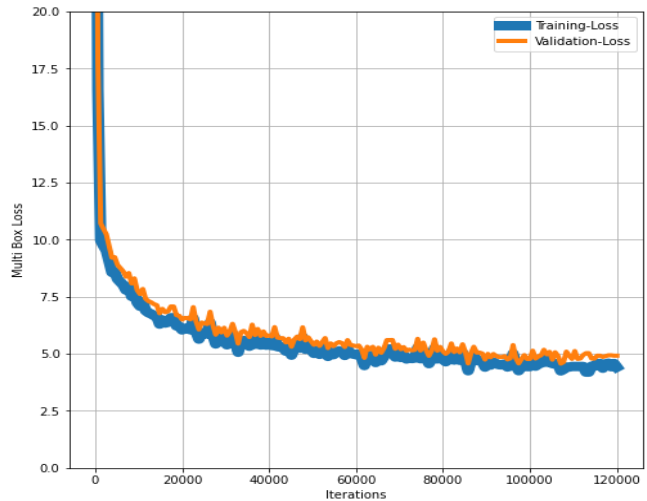

**Fig. 8** Training Loss and Validation Loss of Multi-Box Detection on the Exclusively Dark dataset results.

To choose the best backbone network and proposed structure, we compared the proposed structure with similar work and chose the best backbone model [51]. As most of the proposed model structure attempts to minimize FLOPS, which leads to the downgrading of the accuracy of the proposed model. However, the proposed structure has reached an acceptable level of accuracy while decreasing the computational cost. Furthermore, the network automatically handles objects of varying sizes by combining predictions from many feature maps with varied resolutions. Our model is quicker than object proposing techniques because it offers the best adjustments for better object shape and feature resampling phases and incorporates all computations in a single network. It is easier for training and to integrate with architecture. The detecting feature provides the best efficiency to approaches that include an additional object proposal phase and is substantially faster, according to experimental results here on VOC and Ex-Dark datasets.

Our model mean average precision (mAP) accuracy based on Training and validation on the Exclusively Dark dataset is shown in Fig. 9.

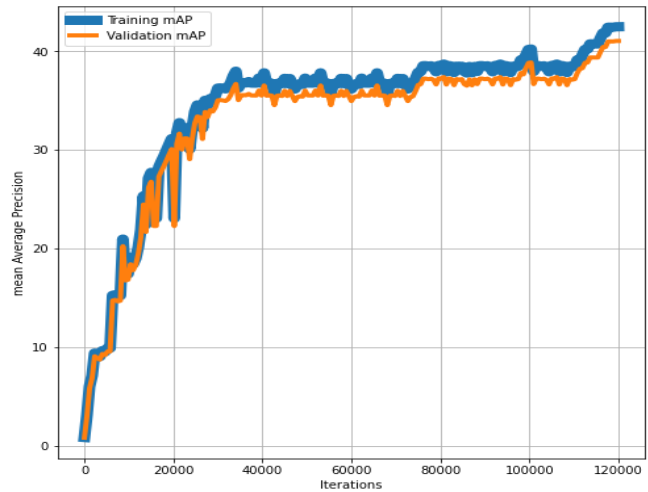

**Fig. 9** Our Model Mean Average Precision mAP accuracy based on Training and Validation on the Exclusively Dark dataset.

Our technique produces state-of-the-art results obtained on the Ex-Dark dataset. Our detecting benchmark is all without frills using the spiking layer in a lighter, efficient system, outperforming all current single-model contributions, such as those in the Ex-Dark datasets. Furthermore, our technique provides a practical and efficient approach to multiscale object detection. In the low light environments, our research frequently addresses the image enhancement challenge, which has little relevance to adaptive systems, and also night vision monitoring, which requires valuable information, but far more relevant themes, such as object recognition are rarely addressed. Results of testing proposed model on different classes of Ex Dark dataset shown in. Fig. 10.

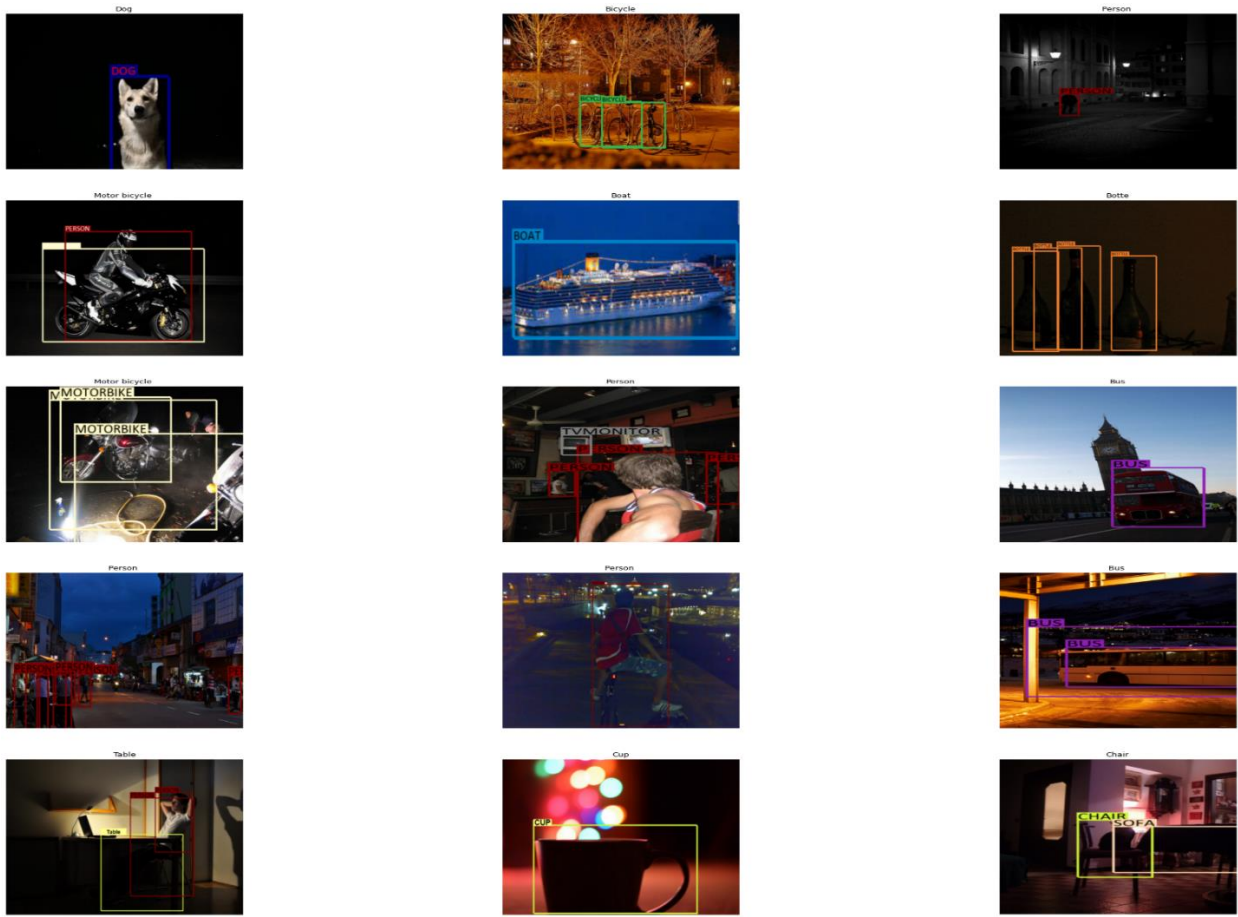

**Fig. 10** Results of the testing proposed model on different classes of Ex Dark dataset.

## 6 Discussion

In this article, a combination of the spiking convolution layer and a conventional convolutional layer is proposed for the detection of dark objects. Proposed structure is composed of two sections. The first sections belong to the backbone of the proposed model that is used as a feature extractor. The second section is used to predict bounding boxes for object detection to decrease the computational cost for the proposed model. We used a spiking convolutional layer in the second section for bounding box detection. With spiking convolutions in the second part, the number of FLOPS for the forward path of the proposed model is 14.9. The proposed structure is further evaluated on the VOC-12 and Ex-dark datasets. The proposed model reached 66.01% and 41.25% mAP for detecting 20 objects in the VOC-12, and 12 objects in the Ex-Dark dataset, respectively. To compare the proposed structure with similar work a comparison between similar works to the proposed model is shown in Fig. 7. Nineteen out of 20 classes have reported better detection results than previous structures. This fact shows the superiority of the proposed model in the case of mean average precision (mAP). Our proposed model is compatible with various versions of CNN model, and the backbone can be changed of the proposed model. This versatility allows the proposed model to be changed based on the need to use the litter model without losing accuracy. Our proposed structure increased the accuracy of structures similar to all versions of tiny YOLO, spiking YOLO, and Faster RCNN by 14% based on the VOC dataset. Results are shown in Fig.11.

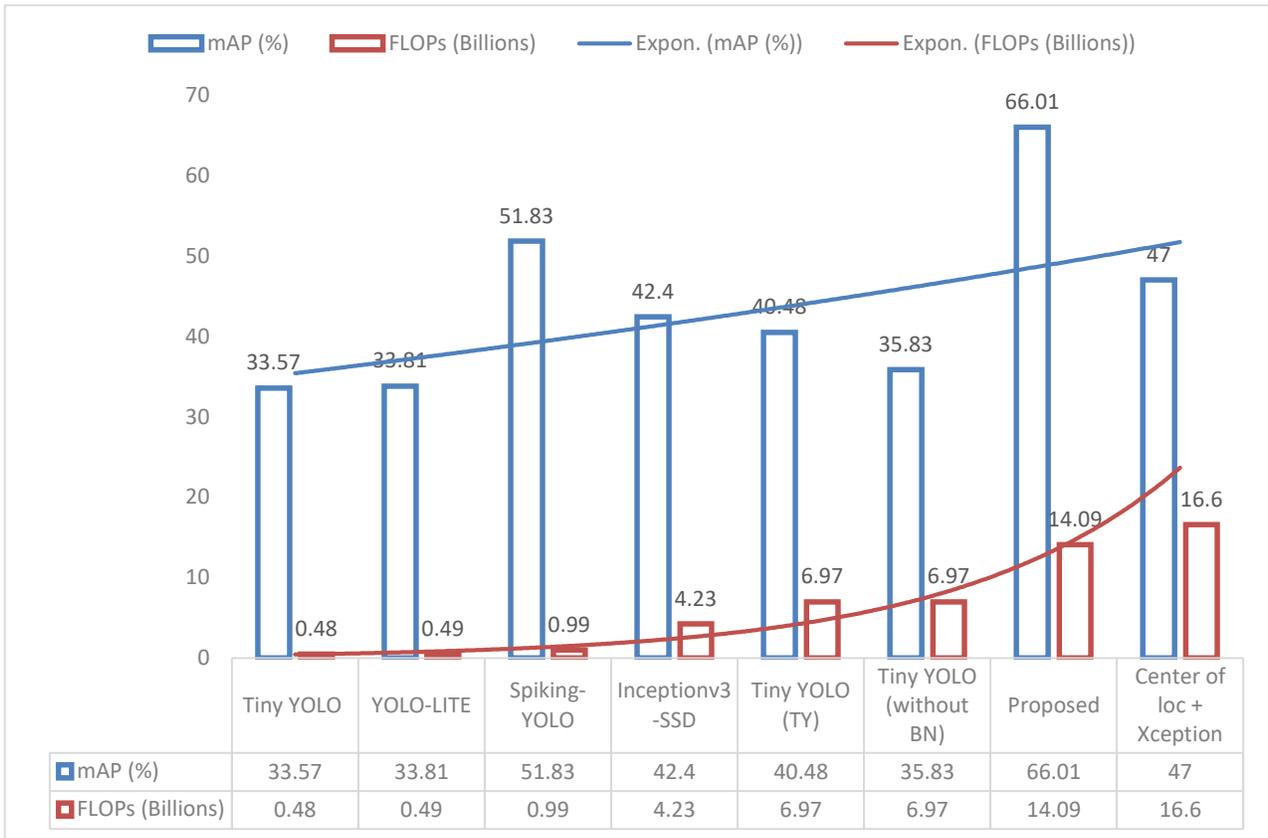

**Fig. 11** Our model Comparison with various object detection models on Ex-dark and VOC dataset [4], [14], [31].

As shown in Fig.11, the mean average precision of our model significantly increases with respect to the number of flops. But our proposed model utilized a Spiking layer with a conventional convolutional layer to minimize and preserve complexity while increasing the number of flops to obtain higher accuracy. Mean Average Precision Object recognition is the most well deep-learning issue wherein algorithms attempt to identify significant items in pictures and classify certain features. The mAP is a key measure that evaluates the accuracy of an object detection model.

## 7 Conclusion

Object detection is a common and ongoing issue in the computer vision field. One of the problems in the field of object detection comes from dark object detection. There is a tradeoff between the state-of-the-art model and its computational cost and complexity to detect dark images because of the low light area. And also, a critical task is to increase the accuracy of the detector model and reduce the computational cost of the model. In this research, a combination of the spiking layer and conventional convolutional layer are proposed to decrease computational complexity while not affecting the overall performance of the proposed model. Our proposed model consisted of the conventional CNN as backbone VGG-16 for feature extraction and a combination of spiking and a conventional convolutional layer for reducing computational cost for detecting dark images. Our proposed model has been evaluated on VOC and Ex-dark datasets. The proposed model reached mAP values of 66.01% and 41.25% for detecting 20 different objects in the VOC-12 dataset and 12 objects in the Ex-dark dataset, respectively. Comparison to similar structures, such as spiking YOLO, Tiny YOLO, LITE YOLO, Tinier YOLO, and the fast RCNN proposed model, reported superior mAP. In the future, evaluating more recently developed Shallow CNN structures to decrease the number of FLOPS while preserving and enhancing accuracy.

## Acknowledgment

This work was supported in part by the National Natural Science Foundation of China under Grant 62133013 and in part by the Chinese Association for Artificial Intelligence (CAAI)-Huawei Mind Spore Open Fund.